\pdfoutput=1

\documentclass[11pt]{article}

\usepackage[final]{coling}

\usepackage{times}
\usepackage{latexsym}

\usepackage[T1]{fontenc}

\usepackage[utf8]{inputenc}

\usepackage{microtype}

\usepackage{inconsolata}

\usepackage{graphicx}

\usepackage{bbm}
\usepackage{amsmath}
\usepackage{amssymb}
\usepackage{hyperref}
\usepackage{graphicx}
\usepackage{booktabs}
\usepackage{multirow}
\usepackage{xcolor}
\usepackage{xspace}
\usepackage{natbib}
\usepackage[most]{tcolorbox}
\usepackage{tabularx}
\usepackage{listings}
\usepackage{mdframed}
\usepackage{subcaption}
\usepackage{textcomp}
\usepackage{algorithm}
\usepackage{algorithmicx}
\usepackage{algpseudocode}

\newcommand{\MAC}{\color[HTML]{3370FF} \textbf{MAC-SQL + GPT-4}}
\newcommand{\ours}{\textbf{\texttt{MAC-SQL}}}

\lstdefinestyle{smallsqlstyle}{language=SQL,basicstyle=\scriptsize\ttfamily,keywordstyle=\color{blue},keepspaces=true}
\lstdefinestyle{sqlstyle}{language=SQL,keywordstyle=\color{blue},keepspaces=true}
\lstdefinestyle{schemastyle}{language={},columns=flexible}

\lstnewenvironment{mytext}{\lstset{language={}}}{}

\newtcolorbox{dialogbox}{
  colback=gray!10!white, colframe=black, sharp corners,
  boxrule=0.5mm, top=10pt, bottom=10pt, left=10pt, right=10pt, breakable
}

\title{MAC-SQL: A Multi-Agent Collaborative Framework for Text-to-SQL}

\author{
Bing Wang\textsuperscript{1}, Changyu Ren\textsuperscript{1}, Jian Yang\textsuperscript{1}, Xinnian Liang\textsuperscript{1}, Jiaqi Bai\textsuperscript{1}, Linzheng Chai\textsuperscript{1} \\
{\bf Zhao Yan\textsuperscript{2}, Qian-Wen Zhang\textsuperscript{2}, Di Yin\textsuperscript{2}, Xing Sun\textsuperscript{2}, Zhoujun Li\textsuperscript{1} \footnotemark[2] }\\ 
\textsuperscript{1}Beihang University  
\textsuperscript{2}Tencent Youtu Lab\\ 
\texttt{\{bingwang,cyren,jiaya,xnliang,bjq,challenging,lizj\}@buaa.edu.cn} \\
\texttt{\{zhaoyan,cowenzhang,endymecyyin,winfredsun\}@tencent.com}\\ }

\begin{document}
\maketitle
\newcommand{\eat}[1]{}

\definecolor{pythonblue}{rgb}{0.16,0.12,0.93}
\definecolor{cppgreen}{rgb}{0.16,0.42,0.16}
\definecolor{promptinsert}{HTML}{bfefff}
\definecolor{codehlcolor}{HTML}{ffec8b}
\definecolor{codehlcolor2}{HTML}{ffbbff}
\definecolor{bgcolor}{rgb}{0.95,0.95,0.92}

\lstdefinestyle{plain}{
    basicstyle=\fontsize{8}{7}\ttfamily,
    keywordstyle=\color{blue},
    commentstyle=\color{gray},
    stringstyle=\color{green},
    showstringspaces=false,
    breaklines=true,
    breakatwhitespace=false,
    breakindent=0pt,
    escapeinside={(*@}{@*)}
}

\lstdefinestyle{sql}{
    language=SQL,
    basicstyle=\fontsize{8}{7}\ttfamily,
    keywordstyle=\color{black},
    commentstyle=\color{black},
    stringstyle=\color{black},
    showstringspaces=false,
    breakatwhitespace=false,
    breaklines=true,
    breakindent=0pt,
    escapeinside={(*@}{@*)}
}

\newcommand{\inserthl}[1]{\sethlcolor{promptinsert}\hl{#1}}
\newcommand{\codehl}[1]{\sethlcolor{codehlcolor}\hl{#1}}
\newcommand{\codehlerr}[1]{\sethlcolor{codehlcolor2}\hl{#1}}

\begin{abstract}
Recent LLM-based Text-to-SQL methods usually suffer from significant performance degradation on ``huge" databases and complex user questions that require multi-step reasoning. Moreover, most existing methods neglect the crucial importance of LLMs using external tools and model collaboration. To address these challenges, we introduce \textbf{\texttt{MAC-SQL}}, a novel LLM-based multi-agent collaborative framework. Our framework comprises a core decomposer agent for Text-to-SQL generation with few-shot chain-of-thought reasoning, accompanied by two auxiliary agents that utilize external tools or models to acquire smaller sub-databases and refine erroneous SQL queries. The decomposer agent collaborates with auxiliary agents, which are activated as needed and can be expanded to accommodate new features or tools for effective Text-to-SQL parsing. In our framework, we initially leverage GPT-4 as the strong backbone LLM for all agent tasks to determine the upper bound of our framework. We then fine-tune an open source instruction-followed model, SQL-Llama, by leveraging Code Llama 7B, to accomplish all tasks as GPT-4 does. Experiments show that SQL-Llama achieves a comparable execution accuracy of 43.94, compared to the baseline accuracy of 46.35 for vanilla GPT-4. At the time of writing, MAC-SQL+GPT-4 achieves an execution accuracy of 59.59 when evaluated on the BIRD benchmark, establishing a new state-of-the-art (SOTA) in its holdout test set.
\footnote{\url{https://github.com/wbbeyourself/MAC-SQL}}
\end{abstract}

\section{Introduction}

\begin{figure}[t]
    \centering
    \includegraphics[width=0.5\textwidth]{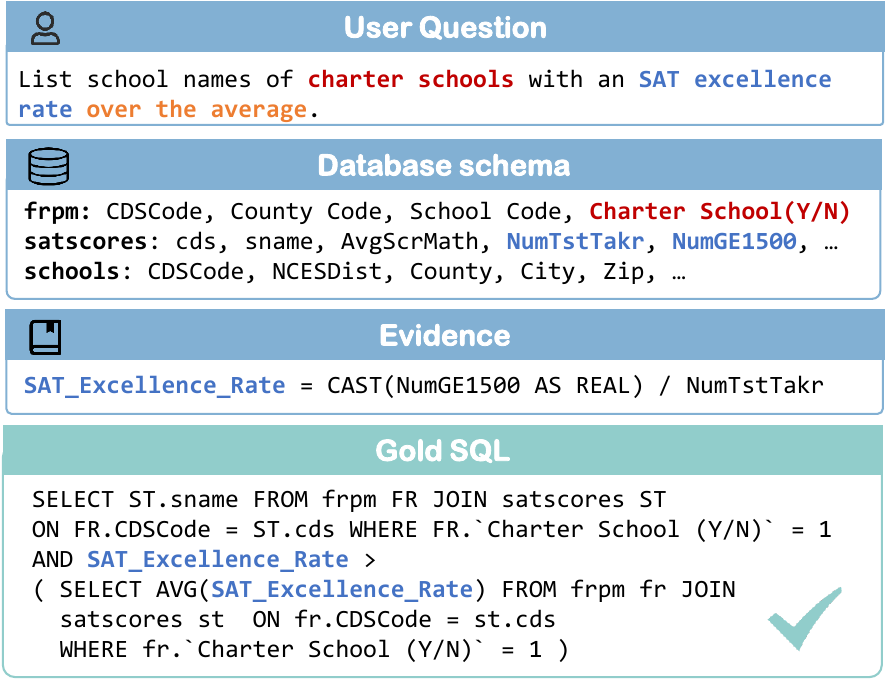}
    \caption{A complex example of Text-to-SQL. In the Gold SQL, we use \texttt{SAT\_Excellence\_Rate} to represent "CAST(NumGE1500 AS REAL)/NumTstTakr" for the sake of brevity.}
    \label{fig:intro-demo}
\end{figure}

Text-to-SQL aims to automate the process of generating Structured Query Language (SQL) queries for databases from natural language text.
This long-standing challenge is essential to improve database accessibility without requiring knowledge of SQL~\citep{qin2022survey,sun2023sqlpalm}.

Over the past decade, research in this field has progressed through three stages. In the initial phase, systems encode the input sequence using pre-trained models, and SQL queries are decoded using either abstract syntax trees~\citep{xu2017sqlnet, guo2019complex, wang2021ratsql} or predefined sketches~\citep{he2019xsql}. More recent systems~\citep{raffel2023exploring,xie2022unifiedskg, scholak2021picard} have adopted sequence-to-sequence methodologies. 
The latest research~\citep{ouyang2022training,OpenAI2023GPT4TR,rozière2023code} has demonstrated the remarkable capabilities of Large Language Models (LLMs) in this task. The success of these models can be attributed to their emerging abilities~\citep{wei2023chainofthought,brown2020language} and the robust reasoning capabilities inherent in LLMs.

\begin{figure*}[ht]
    \centering
    \includegraphics[width=0.99\textwidth]{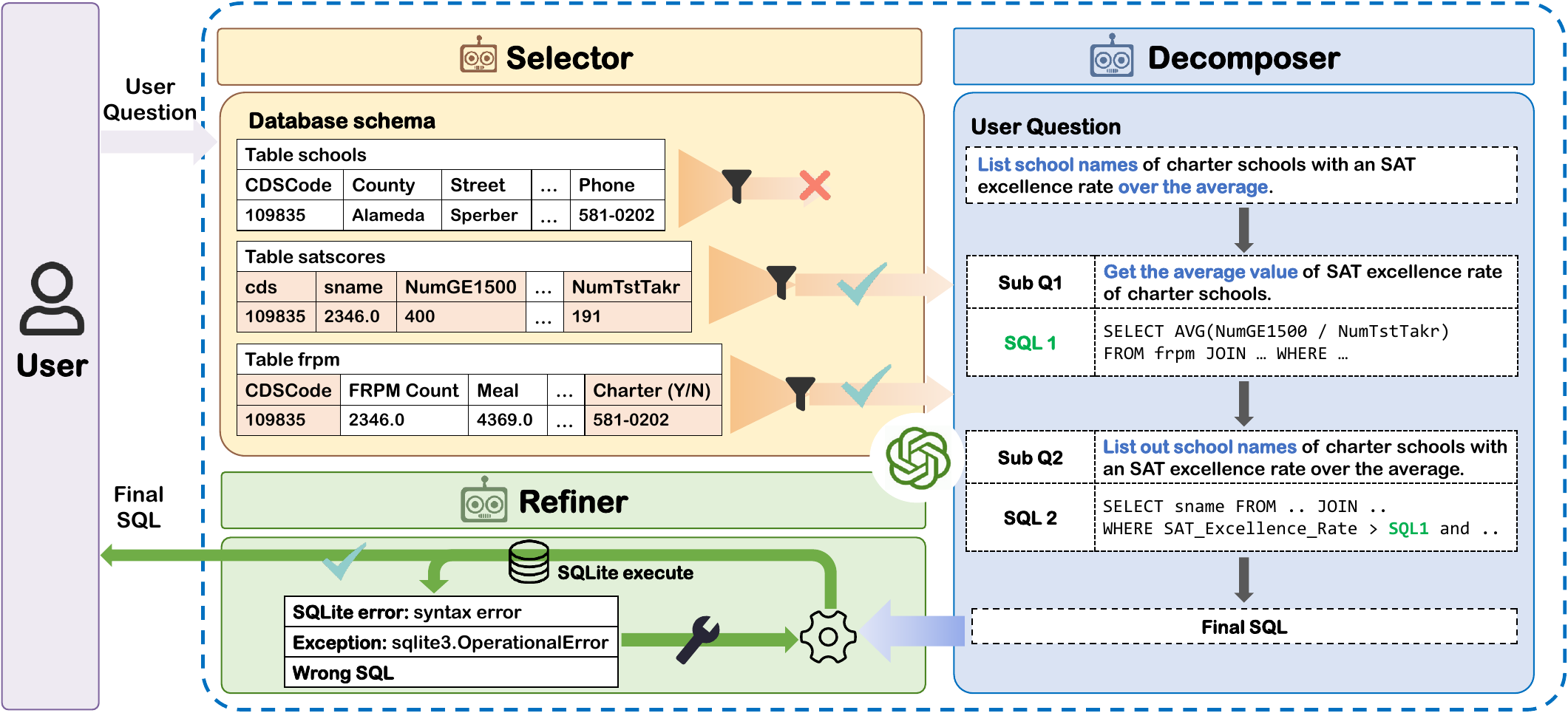}
    \caption{The overview of our \ours{} framework, which comprises three agents: (i) the \textit{Selector}, which decomposes a large database into a smaller sub-database to mitigate the interference of irrelevant information, and (ii) the \textit{Decomposer}, which breaks down a complex question into simpler sub-questions and resolves them progressively by chain-of-thought reasoning, and (iii) the \textit{Refiner}, which uses an external tool for SQL execution and obtains feedback, then refines faulty SQL queries.}
    \label{fig:framework}
\end{figure*}

Recent research on LLM-based Text-to-SQL~\citep{dong2023c3,pourreza2023dinsql,gao2023texttosql} has mainly concentrated on In-Context Learning prompt strategies and supervised fine-tuning using data derived from the target domain. 
However, these approaches usually suffer from significant performance degradation in “huge” databases and complex user questions that require multi-step reasoning, as demonstrated in Figure~\ref{fig:intro-demo}. 
Moreover, most existing methods neglect the crucial importance of LLMs utilizing external tools and model collaboration.

To alleviate the above challenges, we introduce \ours{}, a novel LLM-based multi-agent collaborative framework, which exploits LLMs as intelligent agents with different functionalities for effective Text-to-SQL parsing. 
Our framework comprises a core \textit{Decomposer} agent for Text-to-SQL generation, accompanied by two auxiliary agents, the \textit{Selector} and the \textit{Refiner}, for tool usage and SQL refinement.
Specifically, the Decomposer breaks down a complex question into simpler sub-questions and resolves them progressively by chain-of-thought reasoning.
When necessary, the Selector decomposes a large database into a smaller sub-database to minimize the interference of irrelevant information, while the Refiner employs an external tool for SQL execution, obtains feedback, and refines erroneous SQL queries.

Furthermore, we have fine-tuned an instruction-followed model, SQL-Llama, by leveraging Code Llama 7B, using agent instruction data from \ours{}, thus enabling capabilities in database simplification, question decomposition, SQL generation, and SQL correction.

In our experiments, we initially leverage GPT-4 as a strong backbone LLM for all agent tasks to determine the upper bound of our \ours{} framework on the widely used BIRD and Spider dataset.
Experimental results demonstrate that MAC-SQL+GPT-4 achieves an execution accuracy of 59.59 on the BIRD holdout test set, establishing a new state-of-the-art (SOTA) at the time of writing.
Furthermore, we utilize SQL-Llama (7B) to perform all tasks such as GPT-4.
Surprisingly, despite SQL-Llama having an order of magnitude fewer parameters than GPT-4, its execution accuracy reaches 43.94, which is remarkably close to the accuracy of GPT-4 (46.35).

\textbf{Contribution} Our main contributions and results are summarized as follows:
\begin{enumerate}
    \item We propose \ours{}, a novel multi-agent collaborative framework for Text-to-SQL, which integrates external tools and facilitates model collaboration to address intricate scenarios.
    \item We introduce an instruction-tuning model, named SQL-Llama, to fill in the gaps in open-source agent-instruction-following models for the task of Text-to-SQL.
    \item Experimental results demonstrate that \ours{} achieves state-of-the-art execution accuracy of 59.59\% on the BIRD test set at the time of writing.
\end{enumerate}

\section{Preliminaries}

\subsection{Problem Definition of Text-to-SQL}

Given a triple $\mathcal{X}$ = $(\mathcal{Q}, \mathcal{S}, \mathcal{K})$, where $\mathcal{Q}$, $\mathcal{S}$ and $\mathcal{K}$ are natural language questions, database schema and external knowledge (optional), the database schema $\mathcal{S}$ is defined as $\{\mathcal{T},\mathcal{C}\}$, where $\mathcal{T}$ represents multiple tables $\{\mathcal{T}_{1},\mathcal{T}_{2},\dots,\mathcal{T}_{|T|}\}$ and $\mathcal{C}$ represents columns $\{\mathcal{C}_{1},\mathcal{C}_{2},\dots,\mathcal{C}_{|C|}\}$. 
The purpose of Text-to-SQL task is to generate the correct SQL $\mathcal{Y}$ corresponding to the question $\mathcal{Q}$.

\subsection{Large Language Model for Text-to-SQL}

The Text-to-SQL task has recently been formulated as a generation task~\cite{dong2023c3,pourreza2023dinsql}, designing appropriate prompts to guide a large language model $\mathcal{M}$ generating SQL queries token-by-token. The generation process can be formulated as follows.

\begin{equation}
P_{\mathcal{M}}(\mathcal{Y} | \mathcal{X}) = \prod_{i=1}^{|\mathcal{Y}|} P_{\mathcal{M}}(\mathcal{Y}_i | \mathcal{Y}_{<i}; \mathcal{X})
\label{eq:baset2s}
\end{equation}

where $\mathcal{Y}{<i}$ is the prefix of the SQL query $\mathcal{Y}$ and $P_{\mathcal{M}}(\mathcal{Y}_i | \cdot)$ is the conditional probability of the $i$-th token in the SQL query $\mathcal{Y}$ given the prefix $\mathcal{Y}_{<i}$ and the triple $\mathcal{X}$ = $(\mathcal{Q}, \mathcal{S}, \mathcal{K})$.

\begin{algorithm}[t]
\caption{The algorithm of MAC-SQL}\label{alg:macsql}
\begin{algorithmic}[1]
\Require \texttt{question q}, \texttt{database db}, \texttt{knowledge kg} 
\Ensure \texttt{sql}

\If{need simplify to \texttt{database}}
    \State \texttt{db} = \textcolor[HTML]{3078BE}{$\mathrm{LLM_{Selector}}$}(\texttt{q}, \texttt{db}, \texttt{kg})
\EndIf

\State \texttt{dbDesc} = \textcolor[HTML]{3078BE}{$\mathrm{getDbRepresenation}$}(\texttt{db}, \texttt{kg})
\State \texttt{subQs}, \texttt{subSQLs} = \textcolor[HTML]{3078BE}{$\mathrm{LLM_{Decomposer}}$}(\texttt{q}, \texttt{dbDesc})

\State \texttt{sql} = \texttt{subSQLs[-1]}
\State \texttt{count} = \texttt{0}
\While{$\texttt{count} < \texttt{maxTryTimes}$}
    \State \texttt{ok}, \texttt{err} = \textcolor[HTML]{3078BE}{$\mathrm{executeAndAnalyze}$}(\texttt{sql}, \texttt{db})
    \If{\texttt{ok}}
        \State \Return \texttt{sql}
    \Else
        \State \texttt{sql} = \textcolor[HTML]{3078BE}{$\mathrm{LLM_{Refiner}}$}(\texttt{q}, \texttt{dbDesc}, \texttt{sql}, \texttt{err})
    \EndIf
\EndWhile
\State \Return \texttt{sql}

\end{algorithmic}
\end{algorithm}

\section{MAC-SQL Framework}
\label{sec:method}

\subsection{Overview}
\label{sec:overview}
In Figure~\ref{fig:framework}, we introduce \ours{}, a novel LLM-based multi-agent collaborative framework, which exploits LLMs as intelligent agents with different functionalities for effective Text-to-SQL parsing. 
\ours{} comprises a core \textit{Decomposer} agent for Text-to-SQL generation, accompanied by two auxiliary agents, the \textit{Selector} and the \textit{Refiner}, for tool usage and SQL refinement.
In Algorithm~\ref{alg:macsql}, we present the collaboration process of three agents in \ours{}. 
In the following section, a detailed introduction of three agents will be presented.

\subsection{Selector}
\label{sec:selector}

Given an input triple $\mathcal{X}$ = $(\mathcal{Q}, \mathcal{S}, \mathcal{K})$, where the schema of the database $\mathcal{S} = \{\mathcal{T},\mathcal{C}\}$, the Selector agent aims to locate the minimal schema $\mathcal{S^{'}} = \{\mathcal{T^{'}},\mathcal{C^{'}}\}$, where $T^{'} \subseteq T$ and $C^{'} \subseteq C$, to answer the question $\mathcal{Q}$ with knowledge $\mathcal{K}$. The function of the Selector agent can be described as:

\begin{equation}
\mathcal{S}^{'} = f_{selector}( \mathcal{Q}, \mathcal{S}, \mathcal{K} | \mathcal{M})
\label{eq:selector}
\end{equation}

where $f_{selector}( \cdot | \mathcal{M})$ denotes the function of the selector prompting the LLM $\mathcal{M}$. 
The motivation behind designing the selector involves primarily two key factors.
Firstly, introducing too many irrelevant schema items in the prompt increases the likelihood of LLM generating irrelevant schema items in the output SQL. 
Secondly, using the complete database schema results in excessive text length, leading to unnecessary API costs, and may exceed the maximum context length of LLM.
It is important to note that the Selector will only be activated when the length of the database schema prompt exceeds the length threshold; otherwise, the original database schema $\mathcal{S}$ will be used for the subsequent process.
More details about agent variables and prompts can be found in the Appendix~\ref{appendix-details}.

\subsection{Decomposer}
\label{sec:decomposer}

\begin{figure}[t]
    \centering
    \includegraphics[width=0.5\textwidth]{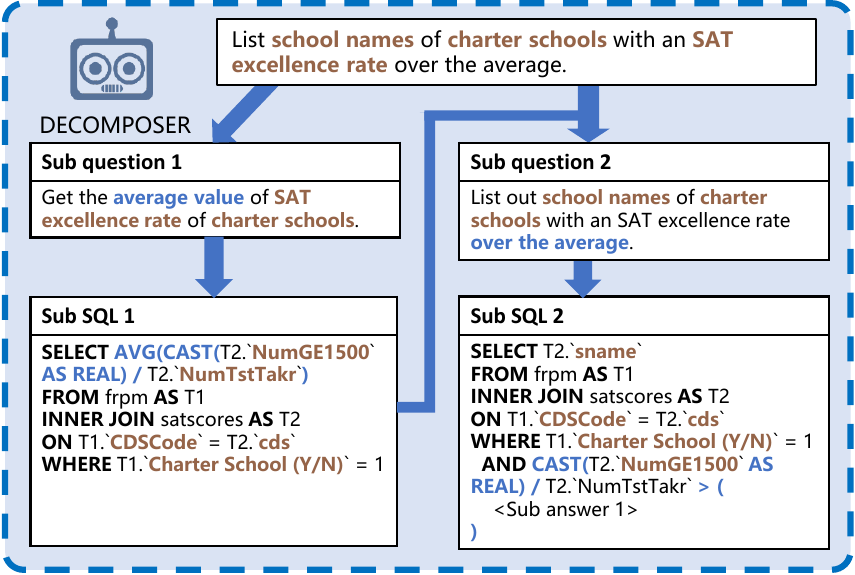}
    \caption{The Decomposer Agent Illustration.}
    \label{fig:decomposer}
\end{figure}

The purpose of the Decomposer is to enhance LLM's reasoning ability by generating a series of intermediate steps (i.e. sub-questions and SQLs) before predicting the final SQL.
As shown in Figure~\ref{fig:decomposer}, the decomposer instructs the LLM to decompose the original complex question $\mathcal{Q}$ as reasoning steps and gets the final SQL query $\mathcal{Y}$ in a single pass.
It can be described as follows.

\begin{equation}
\small
P_{\mathcal{M}}(\mathcal{Y} | \mathcal{Q}, \mathcal{S}^{'}, \mathcal{K}) = \prod_{j=1}^{L} P_{\mathcal{M}}(\mathcal{Y}^{j} | \mathcal{Y}^{<j}; \mathcal{Q}^{j}, \mathcal{S}^{'}, \mathcal{K})
\label{eq:decomposer}
\end{equation}

where $\mathcal{Q}^{j}$ and $\mathcal{Y}^{j}$ are the $j$-th sub-question and sub-SQL generated by the LLM $\mathcal{M}$ given the previous sub-SQLs $\mathcal{Y}^{<j}$, the filtered database schema $\mathcal{S}^{'}$ and knowledge $\mathcal{K}$, $L$ is the number of sub-questions.

The decomposer pattern can be approached in two prompting methods for text-to-SQL parsing: chain-of-thought (CoT) prompting~\citep{wei2023chainofthought} and least-to-most prompting~\citep{zhou2022least-to-most}.
The former involves generating thinking and reasoning once to obtain an answer, while the latter brings about higher computational costs to generate each SQL query due to the iterative process.

Due to the inefficiency of the iterative method and the necessity to determine the stopping criteria, we adopt the CoT approach to generate sub-questions and their corresponding SQL queries.
The specific implementation is as follows: dynamically judging the difficulty of the user's question, if it can be answered by a simple SQL query, then the SQL is generated directly. 
If the question is more complex, the corresponding SQL is generated starting from the simplest sub-question, and then gradually broken down to obtain progressive sub-questions until the final SQL corresponding to the question is obtained. 
Additionally, we leverage the few-shot approach to enhance LLM's understanding of instructions through in-context learning.

\subsection{Refiner}
\label{sec:refiner}

\begin{figure}[t]
    \centering
    \includegraphics[width=0.5\textwidth]{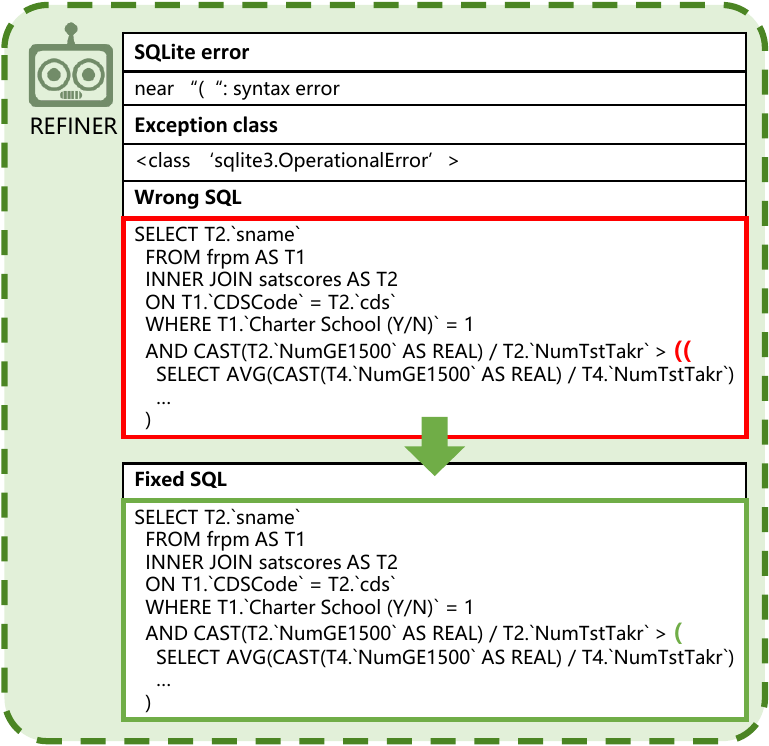}
    \caption{The Refiner Agent Illustration.}
    \label{fig:refiner}
\end{figure}

The primary function of the Refiner is to detect and automatically correct SQL errors, as illustrated in Figure~\ref{fig:refiner}. 
In a comprehensive multi-agent collaborative framework, particularly within the context of Text-to-SQL tasks, the refiner is essential for the inspection and correction of generated answers. 
For instance, in the ChatDev project~\cite{qian2024chatdev}, intelligent agents are responsible for conducting overall and functional module testing in addition to overall architectural design and code writing for game software development tasks. 
Similarly, in Text-to-SQL tasks, the Refiner can be used to make appropriate adjustments for the different datasets, database schemas, SQL generation styles, and specific inductive biases.

Given a flawed SQL query $\mathcal{Y}^{'}$ and the error message feedback $\mathcal{E}$, obtained from external SQL tools, the Refiner instructs the LLM $\mathcal{M}$ to generate the correct SQL query $\mathcal{Y}$.
It can be described as follows.

\begin{equation}
\mathcal{Y} = f_{refiner}( \mathcal{E}, \mathcal{Y}^{'}, \mathcal{Q}, \mathcal{S}^{'}, \mathcal{K} | \mathcal{M})
\label{eq:refiner}
\end{equation}

where $f_{refiner}( \cdot | \mathcal{M})$ denotes the function of the Refiner by prompting the LLM $\mathcal{M}$.

As shown in Figure~\ref{fig:framework}, upon receiving an SQL query, the Refiner diagnoses the SQL statement to assess its syntactic correctness, execution feasibility, and retrieval of non-empty results from the database. 
In general, the purpose of the Refiner is to achieve self-checking and self-correction of the model to enhance the overall framework's fault tolerance and accuracy.
Using the refiner agent, there is a significant reduction in syntax errors, schema linking, and other simple errors. 

\section{SQL-Llama Model}

\subsection{Instruction Dataset Construction}

To construct the Agent-Instruct dataset, we instruct the GPT-4 with the training set of the BIRD and Spider dataset through multi-agent tasks. 
We collect the generated instruction data according to the level of difficulty and filter out those with incorrect SQL query output.
Finally, the curated Agent-Instruct dataset $\mathcal{D}$ with instruction tasks $N$ (N = 3), $\mathcal{D} = \{\mathcal{D}_{i}\}_{i=1}^{N}$ contains 10,000 high-quality instruction data with 3 agent-instruction tasks, covering the distribution of the BIRD and Spider dataset.

\subsection{Multi-task Supervised Fine-tuning}

Our research has primarily focused on the development of open source models within the \ours{} framework, to achieve performance levels comparable to closed source models such as GPT-4. 
To achieve this, we have put significant effort into preparing the data for model training and have open-sourced SQL-Llama, a model that has been fine-tuned using three intelligent agent instruction data. The SQL-Llama model, based on Code Llama 7B, has undergone supervised fine-tuning using agent instruction data from \ours{}, which has enhanced its capabilities in database simplification, question decomposition, SQL generation, and SQL correction.

Given the Agent-Instruct dataset with instruction tasks $N$ (N = 3), $\mathcal{D} = \{\mathcal{D}_{i}\}_{i=1}^{N}$, the LLM trained in $D$ can learn from these tasks and complete agent tasks. 
The supervised fine-tuning process can be described as:

\begin{equation}
\small
\mathcal{L} = -\sum_{i=1}^{N} \mathbb{E}_{\mathcal{Q},\mathcal{S}^{i},\mathcal{K},\mathcal{Y}^{i} \sim \mathcal{D} } \left[ \log P(\mathcal{Y}^{i} | \mathcal{Q}, \mathcal{S}^{i}, \mathcal{K}; \mathcal{M}) \right]
\label{eq:sft}
\end{equation}

where $\mathcal{L}$ is the training objective of $N$ tasks, $\mathcal{S}^{i}$ and $\mathcal{Y}^{i}$ are the selected database schema and the intermediate SQL query of the $i$-th task.

One of the key challenges we encountered during the model training process was balancing model complexity with performance. We had to carefully optimize the model architecture and parameters to ensure that it could effectively handle the complexities of database-related tasks while still maintaining high-performance levels. Additionally, ensuring the quality and relevance of the instruction dataset for training was crucial, as it directly impacted the model's performance.

Despite these challenges, our work on instruction-tuned models represents a significant step towards democratizing access to high-performance language models for database-related tasks. By open-sourcing both the model and the instruction dataset, we aim to provide valuable resources for further research and development in this area, ultimately leading to more accessible and effective tools for database query processing and related tasks.

\section{Experiments}

\subsection{Experimental Setup}
\label{experiment_setup}

\paragraph{Datasets} 
The Spider~\citep{YuSpider2018} dataset is frequently used to assess the performance of text-to-SQL parsing across multiple databases, which requires models to demonstrate adaptability to unfamiliar database structures. 
The dataset comprises 7,000 question-query pairs in the training set and 1,034 pairs in the development set, covering 200 distinct databases and 138 domains.

The BIRD~\citep{li2023llm} dataset released by Alibaba DAMO Academy is a new benchmark for real large-scale databases, containing 95 large-scale databases and high-quality Text-SQL pairs, with a data storage volume of up to 33.4GB covering 37 professional domains. Unlike Spider, BIRD focuses on massive and real database content, external knowledge reasoning between natural language questions and database content, and new challenges in SQL efficiency when dealing with large databases.

\paragraph{Evaluation Metrics}

Following BIRD~\citep{li2023llm} and Test-suite~\citep{zhong-etal-2020-semantic}, we consider three metrics, exact match accuracy (EM), execution accuracy (EX), and valid efficiency score (VES) to evaluate text-to-SQL models faced with real-world scenarios with large database contents. 
\textit{Exact Match Accuracy (EM)} treats each clause as a set and compares the prediction for each clause with its corresponding clause in the reference query. A predicted SQL query is considered correct only if all of its components match the ground truth. This metric does not take values into account. 
\textit{Execution Accuracy (EX)} is defined as the proportion of questions in the evaluation set for which the execution results of both the predicted and ground truth inquiries are identical, relative to the total number of queries. 
\textit{Valid Efficiency Score (VES)} is designed to measure the efficiency of valid SQLs generated by models. It is important to note that "valid SQLs" refers to predicted SQL queries whose result sets align with those of the ground-truth SQLs.

\paragraph{Baselines} We conduct experiments on both BIRD and Spider datasets and compare our method with the following baseline:
\begin{itemize}
    \item \textbf{GPT-4}~\citep{OpenAI2023GPT4TR} uses a simple zero-shot text-to-SQL prompt for SQL generation.
    \item \textbf{DIN-SQL}~\citep{pourreza2023dinsql} decomposes the text-to-SQL task into smaller subtasks and designs different prompts for each subtask to instruct GPT-4  to complete each subtask and obtain the final SQL.
    \item \textbf{DAIL-SQL}~\citep{gao2023texttosql} encodes structure knowledge as SQL statements, selects few-shot demonstrations based on their skeleton similarities, and removes cross-domain knowledge from examples for token efficiency.
    \item \textbf{C3-SQL}~\citep{dong2023c3} first performs schema linking filtering and then directs GPT-4 with a calibration bias prompt designed for Spider using a self-consistency strategy.
\end{itemize}

\subsection{Overall Performance}

\begin{table}[]
\small
\centering
\begin{tabular}{@{}lcccc@{}}
\toprule
  & \multicolumn{2}{c}{\textbf{Dev}} & \multicolumn{2}{c}{\textbf{Test}} \\ \cmidrule(l){2-5} 
\multirow{-2}{*}{\textbf{Method}}               & \textbf{EX}    & \textbf{VES}    & \textbf{EX}     & \textbf{VES}    \\ 

\midrule

Palm-2                                          & 27.38          & -               & 33.04           & -              \\
ChatGPT + CoT                                    & 36.64          & 42.30           & 40.08           & 56.56          \\
Claude-2                                        & 42.70          & -               & 49.02           & -              \\
GPT-4                                           & 46.35          & 49.77           & 54.89           & 60.77           \\
DIN-SQL + GPT-4                                 & 50.72          & 58.79           & 55.90           & 59.44           \\
DAIL-SQL + GPT-4                                & 54.76          & 56.08           & 57.41           & 61.95           \\  

\midrule

SQL-Llama                     & 32.87          & 55.67           & -  & -  \\ 
\ours{} + SQL-Llama           & 43.94          & 57.36           & -  & -  \\ 
\quad + Oracle Schema         & 51.43          & 58.24           & -  & -  \\ 
\ours{}+GPT-3.5-Turbo       & 50.56          & 61.25           & -  & -  \\ 
\quad + Oracle Schema         & 65.78         & 60.62           & -  & -  \\ 
\MAC                  & \textbf{59.39}  & \textbf{66.39} & \textbf{59.59}  & \textbf{67.68}  \\ 
\quad + Oracle Schema         & 70.28          & 62.63           & -  & -  \\

\bottomrule
\end{tabular}
\caption{Execution accuracy(EX) and Valid efficiency score (VES) on both dev and test set of BIRD dataset. The SQL-Llama model refers to version 7B. The term "Oracle Schema" refers to the utilization of a ground truth sub-database as the input for the Decomposer, rather than employing the results obtained from the Selector.}
\label{tab:bird-main-result}
\end{table}

It is important to note that the experiment utilized the 32k version of GPT-4 and the 16k version of GPT-3.5-Turbo.
\paragraph{BIRD Results} In Table~\ref{tab:bird-main-result}, we report the performance of our method and baseline methods on the BIRD dataset.
It is evident that our method surpasses all LLM-based methods in terms of execution accuracy (EX) and valid efficiency score (VES) in both the development and test sets. 
Specifically, our method outperforms the second-best method by 4. 63\% in the development set and by 2. 18\% in the test set. 
At the time of writing, \ours{}+GPT-4 achieves an execution accuracy of 59.59 when evaluated on the BIRD benchmark, establishing a new state-of-the-art (SOTA) in its holdout test set.

\paragraph{Spider Results} 
Currently, Spider has open-sourced the test set, so we can evaluate our method in both the development and the test set.
As shown in Table~\ref{tab:spider-main-result}, for the Spider dev set~\citep{YuSpider2018}, our method achieves the highest execution accuracy using GPT-4. 
These results demonstrate the generalization ability of our \ours{} framework.

\begin{table}[]
\centering
\begin{tabular}{lcc}
\toprule
\textbf{Method}        & \textbf{Dev}   & \textbf{Test} \\
\midrule
C3 + ChatGPT              & 81.80           & 82.30                  \\
DIN-SQL + GPT-4           & 82.80           & 85.30                  \\
DAIL-SQL + GPT-4          & 84.40           & \textbf{86.60}         \\

\midrule

SQL-Llama              & 65.48          & 61.63           \\ 
\ours{}+SQL-Llama    & 76.25          & 70.58           \\
\ours{}+GPT-3.5-Turbo   & 80.56          & 75.53           \\
\MAC                      & \textbf{86.75} & 82.80           \\

\bottomrule
\end{tabular}
\caption{Execution accuracy(EX) on both dev and test set of Spider. The SQL-Llama model refers to version 7B.}
\label{tab:spider-main-result}
\end{table}

\begin{table}[t]
\small
\centering
\begin{tabularx}{\columnwidth}{@{}l*{4}{X}@{}}
\toprule
\textbf{Method}                                 & \textbf{Simple} & \textbf{Mod.} & \textbf{Chall.} & \textbf{All}   \\ \midrule
{\color[HTML]{3370FF} \textbf{MAC-SQL+GPT-4}} & 65.73           & 52.69             & 40.28                & \textbf{59.39} \\
\quad w/o Selector                                    & 65.73           & 52.04             & 35.14                & 57.28   \\
\quad w/o Decomposer                                  & 61.51           & 48.82             & 38.89                & 55.54    \\
\quad w/o Refiner                                     & 63.24           & 44.52             & 33.33                & 54.76   \\ \bottomrule
\end{tabularx}
\caption{Execution accuracy of \ours{} ablation study in BIRD dev set. For brevity, the abbreviation "Mod." stands for "Moderate" while "Chall." denotes "Challenging".}
\label{tab:bird-ablation}
\end{table}

\subsection{Ablation Study}

Table~\ref{tab:bird-ablation} presents the results of an ablation study for the \ours{} model in the BIRD dev set. 
The table lists different variations of the \ours{} model, including with and without certain components such as Selector, Decomposer, and Refiner. 
The other columns represent the accuracy of the model at different levels of difficulty: Simple, Moderate, and Challenging, as well as the overall accuracy (All).

The findings show that the original \ours{} + GPT-4 model achieves an accuracy of 65. 73\% in Simple, 52. 69\% in Moderate and 40. 28\% in Challenging, with an overall accuracy of 59.39\%. 
When removing the Selector component, the accuracy remained the same for Simple, but decreased to 52.04\% for Moderate and 35.14\% for Challenging, resulting in an overall accuracy of 57.28\% (a decrease of 2.11\%). 
Similarly, removing the Decomposer and Refiner components also led to decreased accuracy across all difficulty levels.

In general, the ablation study indicates that each component of the \ours{} model (Selector, Decomposer, and Refiner) plays a crucial role in achieving high accuracy, as their removal resulted in decreased performance at all difficulty levels.

\subsection{Discussion}

\paragraph{Impact on the number of demonstrations} 
Table~\ref{tab:few-shot} shows the evaluation results of \ours{} with different numbers of demonstrations in the BIRD and Spider datasets. 
As the number of shots increases from 0 to 2, there is a consistent improvement in the performance metrics (EX, VES, and EM) for both BIRD and Spider. 
This indicates that the model benefits from additional demonstration examples and is able to generalize better with more data. 
The highest performance is achieved with 2-shot evaluation, indicating that the model is capable of effectively learning from a small number of examples.
The high cost of the GPT-4 interface results in a significant consumption of tokens during a full test of the dev set for Spider and BIRD, estimated at approximately 6 million and 10 million tokens, respectively.
Due to cost constraints, our analysis is limited to a maximum of 2 shots, and further experiments involving more shots (e.g., shot $k > 2$) will have to await a more budget-friendly implementation of GPT-4.

\begin{table}[]
\centering
\begin{tabular}{@{}ccccc@{}}
\toprule
\multicolumn{1}{c}{\multirow{2}{*}{\textbf{Few-shot}}} & \multicolumn{2}{c}{\textbf{BIRD}} & \multicolumn{2}{c}{\textbf{Spider}} \\ \cmidrule(l){2-5} 
\multicolumn{1}{c}{}                                   & \textbf{EX}     & \textbf{VES}    & \textbf{EM}      & \textbf{EX}      \\ \cmidrule(r){1-5}
0-shot                                                 & 55.54           & 63.31           & 58.42            & 74.22            \\
1-shot                                                 & 57.26           & 64.32           & 59.68            & 78.35            \\
2-shot                                                 & \textbf{59.39}  & \textbf{66.24}  & \textbf{63.20}   & \textbf{86.75}   \\ \bottomrule
\end{tabular}
\caption{Results of \ours{}+GPT-4 on the dev set of BIRD and Spider with few-shot evaluation.}
\label{tab:few-shot}
\end{table}

\subsection{Error Analysis}

\begin{figure*}[t]
    \centering
    \includegraphics[width=1.0\textwidth]{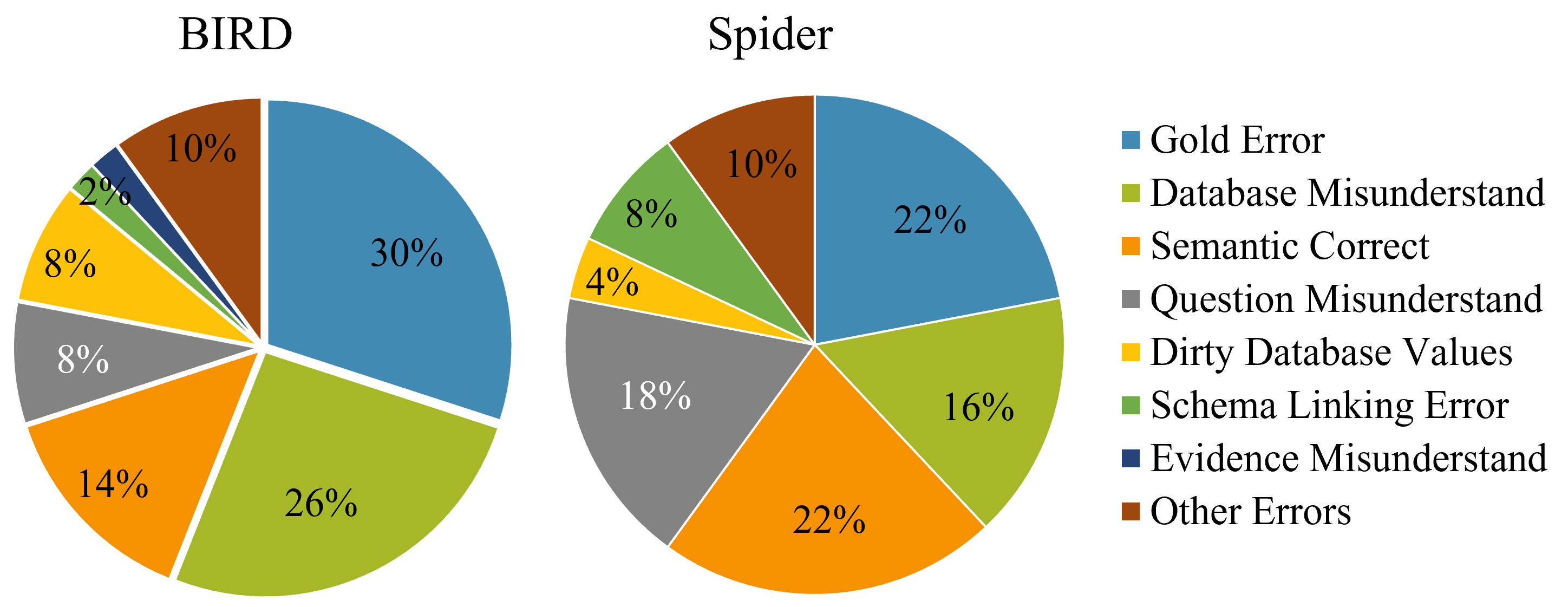}
    \caption{Error Distributions of \ours{} on dev set of BIRD and Spider.}
    \label{fig:error_distribution}
\end{figure*}

In order to thoroughly assess the limitations of our method, we begin by choosing two datasets (BIRD and Spider) that contain various types of structured data, as shown in Figure~\ref{fig:error_distribution}. 

Figure~\ref{fig:error_distribution} displays the error type distribution in the BIRD and Spider datasets. "Gold Error" is the most common error type, accounting for 30\% and 22\% in BIRD and Spider, respectively, signifying the significance of gold standard annotations. "Semantic Correct" is another prevalent error type, representing 14\% and 22\% in BIRD and Spider, respectively, indicating the importance of semantic understanding and correctness. However, the "Schema Linking Error" is more frequent in BIRD (2\%) than in Spider (8\%), demonstrating differences in schema linking errors.
This analysis underscores the need for addressing gold standard annotations, semantic correctness, and schema linking in dataset development and evaluation, thereby improving their quality and reliability. 
The Appendix \ref{sec:error-examples} contains detailed examples of error types.

\section{Related Work}

\paragraph{LLMs for Text-to-SQL} 

Recent advancements in text-to-SQL tasks using large language models (LLMs) have focused on improving prompt design and developing multi-stage refined frameworks. 
In the early stages of the emergence of large language models, research efforts primarily focused on designing high-quality prompts to better exploit the potential of LLMs for SQL generation. 
For example, \citep{tai2023exploring} systematically studied how to enhance LLM's reasoning ability through chain-of-thought style prompting, including the original chain-of-thought prompting and least-to-most prompting. 
Similarly, ~\citep{chang2023prompt} comprehensively investigated the impact of prompt constructions in various settings when constructing the prompt text for text-to-SQL input. 
Furthermore, DAIL-SQL~\citep{gao2023texttosql} systematically examined prompt engineering for LLM-based Text-to-SQL methods, including question representations, prompt components, example selections, and example organizations.
Later studies, such as C3-SQL~\citep{dong2023c3}, DIN-SQL~\citep{pourreza2023dinsql}, and StructGPT~\citep{jiang2023structgpt}, proposed frameworks for simplifying databases, generating SQL, verifying queries, and integrating answers through zero-shot approaches, query decomposition, and specialized interfaces for structured data access.

However, the aforementioned methods have several issues. 
Firstly, the experiments were conducted solely on the Spider family dataset, failing to demonstrate their generalization to more complex datasets like BIRD, hence limiting their real-world applicability. 
Secondly, certain methods depend on difficulty-level classifiers and customized biases specific to the Spider dataset for error correction, thus lacking the ability to generalize to a broader spectrum of error types.
Third, these methods neglect the utilization of external tools and the collaboration of different modules.
Thus, we propose a framework centered on multi-agent collaboration that can be utilized for more intricate data scenarios and a broader spectrum of error types for detection and correction.

\paragraph{LLM-based Agents} LLM-based agents have been a prominent area of study in both the academic and industry communities for an extended period~\citep{wang2023survey}. Recently, through the acquisition of vast amounts of web knowledge, LLMs have demonstrated remarkable potential in achieving human-level intelligence. This development has led to a surge in research exploring LLM-based autonomous agents. AutoGPT~\citep{autogpt2023} is an open-source implementation of an AI agent and follows a single-agent paradigm in which it augments the AI model with many useful tools, and does not support multi-agent collaboration. Similarly, OpenAgents~\citep{OpenAgents} develops three distinct agents, the Data Agent for data analysis, the Plugins Agent for plugin integration, and the Web Agent for autonomous web browsing, each specializing in different domains, similar to OpenAI's ChatGPT Plugins. Additionally, AutoGen~\citep{wu2023autogen} is an open source framework that enables developers to build customizable, conversable agents that can operate in various modes, employing combinations of LLMs, human input, and tools to perform tasks. However, how to apply LLM-based agents to Text-to-SQL parsing remains under-explored.

While previous studies have focused on single-agent paradigms or domain-specific applications, there is a lack of research on multi-agent collaborative frameworks for Text-to-SQL parsing. We aim to address this gap by proposing a novel approach that integrates multiple LLM-based agents to collectively interpret SQL queries. By leveraging the strengths of different agents specialized in various aspects of Text-to-SQL parsing, our framework aims to improve the accuracy and efficiency of SQL query interpretation in real-world scenarios.

\section{Conclusion}
In summary, this paper proposes the \ours{} framework, which utilizes multi-agent collaboration to address challenges in Text-to-SQL tasks. 
The framework, along with the open source SQL-Llama model, achieved an execution accuracy of 59.59 when evaluated on the BIRD benchmark, establishing a new state-of-the-art (SOTA) on its holdout test set. 
This work presents a novel approach to Text-to-SQL and provides practical guidance to achieve high performance in this domain.
Furthermore, our framework can be expanded to support a broader spectrum of scenarios.

\section*{Limitations}
The agent prompts utilized in the work may benefit from further optimization and might not represent the most optimal choice.
Furthermore, this paper reports the fine-tuning results of the 7B CodeLLama model. Although it performs at a comparable level, we believe its performance can be further improved by using larger models.

\section*{Ethics Statement}
The datasets and models utilized in this paper, and the implementation of the code and the resulting models, are not associated with any ethical concerns.

\section*{Acknowledgments}

This work was partially supported by the National Natural Science Foundation of China (Grant Nos. 62276017, 62406033, U1636211, 61672081), and the State Key Laboratory of Complex\& Critical Software Environment (Grant No. SKLCCSE-2024ZX-18).

\bibliography{custom}

\appendix
\onecolumn

\section{Implementation Details}
\label{appendix-details}

\subsection{Selector Agent}

The selector Agent is activated only when encountering large databases. In the specific implementation, there are two methods to determine whether the current database is large. The first method involves calculating the token count of the database schema string, such as len(tokens) > (0.8 * max\_sequence\_length\_of\_model) (for example, in this study, using GPT-4-32k, len(tokens) > 25k is considered a large database); the second method involves counting the total number of columns and the average number of columns in the tables, which can be adjusted based on the situation. The experimental results in this study are obtained using the first method. For a specific code implementation, the \_is\_need\_prune function in agents.py can be modified accordingly. In the design of the selector prompt, in order to guide the model to output in the specified JSON format, we provide a one-shot example, as detailed in the appendix \ref{appendix-selector-prompt}. After the database is filtered by the selector, to ensure completeness, each table will retain at least 6 column names, preventing missing columns.

\subsection{Decomposer Agent}

The decomposer utilizes a maximum of two shots, as outlined in the appendix \ref{appendix-decomposer-prompt}. The database schema comprises the database ID, table name, column name, full column name/description, and high-frequency cell values. When examining cell values, we consider the column type, unique values, and maximum length. Columns containing only numerical values, as well as unconventional values, such as URLs and emails, are ignored. The decomposer breaks down the original question into multiple sub-questions, ranging from one to five. In the case of a single sub-question, it denotes a straightforward question that can be directly solved in one step without further decomposition into finer ones.

\subsection{Refiner Agent}

The refiner is tasked with rectifying problematic SQL queries. If the SQL query contains multiple issues, it may require multiple correction processes. To ensure practical efficiency, a maximum of three rounds of error correction will be performed. Common error types include SQL syntax errors, schema illusions (such as non-existent table and column names), and empty query results. Presently, the limitation of the refiner is that in cases where the SQL query runs without error and generates non-empty results, even if the SQL does not align with the intended question, the refiner will not make further corrections. We will continue to investigate how to address such ambiguous scenarios in future research.

\section{Prompt Details}
\label{appendix-prompt}

\subsection{Selector Prompt}
\label{appendix-selector-prompt}

\begin{dialogbox}
\begingroup
As an experienced and professional database administrator, your task is to analyze a user question and a database schema to provide relevant information. The database schema consists of table descriptions, each containing multiple column descriptions. Your goal is to identify the relevant tables and columns based on the user question and evidence provided. \\

\textbf{[Instruction]} \\
1. Discard any table schema that is not related to the user question and evidence. \\
2. Sort the columns in each relevant table in descending order of relevance and keep the top 6 columns. \\
3. Ensure that at least 3 tables are included in the final output JSON. \\
4. The output should be in JSON format. \\

\textbf{[Requirements]} \\
1. If a table has less than or equal to 10 columns, mark it as "keep\_all". \\
2. If a table is completely irrelevant to the user question and evidence, mark it as "drop\_all". \\
3. Prioritize the columns in each relevant table based on their relevance. \\

Here is a typical example: \\

========== \\
\text{[}DB\_ID\text{]} banking\_system \\
\text{[}Schema\text{]} \\
\# Table: account \\
\text{[} \\
\text{\ \ \ \ } (account\_id, the id of the account. Value examples: \text{[}11382, 11362, 2, 1, 2367].), \\
\text{\ \ \ \ } (district\_id, location of branch. Value examples: \text{[}77, 76, 2, 1, 39].), \\
\text{\ \ \ \ }  (frequency, frequency of the acount. Value examples: \text{[}'POPLATEK MESICNE', 'POPLATEK TYDNE', 'POPLATEK PO OBRATU'].), \\
\text{\ \ \ \ }  (date, the creation date of the account. Value examples: \text{[}'1997-12-29', '1997-12-28'].) \\
\text{]}  \\
\# Table: client \\
\text{[}  \\
\text{\ \ \ \ }  (client\_id, the unique number. Value examples: \text{[}13998, 13971, 2, 1, 2839\text{]}.),  \\
\text{\ \ \ \ }  (gender, gender. Value examples: \text{[}'M', 'F'\text{]}. And F:female . M:male ),  \\
\text{\ \ \ \ }  (birth\_date, birth date. Value examples: \text{[}'1987-09-27', '1986-08-13'\text{]}.),
\text{\ \ \ \ }  (district\_id, location of branch. Value examples: \text{[}77, 76, 2, 1, 39\text{]}.)  \\
\text{]} \\
\# Table: loan \\
\text{[} \\
\text{\ \ \ \ }  (loan\_id, the id number identifying the loan data. Value examples: \text{[}4959, 4960, 4961\text{]}.), \\
\text{\ \ \ \ }  (account\_id, the id number identifying the account. Value examples: \text{[}10, 80, 55, 43\text{]}.), \\
\text{\ \ \ \ }  (date, the date when the loan is approved. Value examples: \text{[}'1998-07-12', '1998-04-19'\text{]}.), \\
\text{\ \ \ \ }  (amount, the id number identifying the loan data. Value examples: \text{[}1567, 7877, 9988\text{]}.), \\
\text{\ \ \ \ }  (duration, the id number identifying the loan data. Value examples: \text{[}60, 48, 24, 12, 36\text{]}.), \\
\text{\ \ \ \ }  (payments, the id number identifying the loan data. Value examples: \text{[}3456, 8972, 9845\text{]}.), \\
\text{\ \ \ \ }  (status, the id number identifying the loan data. Value examples: \text{[}'C', 'A', 'D', 'B'\text{]}.) \\
\text{]} \\
\# Table: district \\
\text{[} \\
\text{\ \ \ \ }  (district\_id, location of branch. Value examples: \text{[}77, 76\text{]}.), \\
\text{\ \ \ \ }  (A2, area in square kilometers. Value examples: \text{[}50.5, 48.9\text{]}.), \\
\text{\ \ \ \ }  (A4, number of inhabitants. Value examples: \text{[}95907, 95616\text{]}.), \\
\text{\ \ \ \ }  (A5, number of households. Value examples: \text{[}35678, 34892\text{]}.), \\
\text{\ \ \ \ }  (A6, literacy rate. Value examples: \text{[}95.6, 92.3, 89.7\text{]}.), \\
\text{\ \ \ \ }  (A7, number of entrepreneurs. Value examples: \text{[}1234, 1456\text{]}.), \\
\text{\ \ \ \ }  (A8, number of cities. Value examples: \text{[}5, 4\text{]}.), \\
\text{\ \ \ \ }  (A9, number of schools. Value examples: \text{[}15, 12, 10\text{]}.), \\
\text{\ \ \ \ }  (A10, number of hospitals. Value examples: \text{[}8, 6, 4\text{]}.), \\
\text{\ \ \ \ }  (A11, average salary. Value examples: \text{[}12541, 11277\text{]}.), \\
\text{\ \ \ \ }  (A12, poverty rate. Value examples: \text{[}12.4, 9.8\text{]}.), \\
\text{\ \ \ \ }  (A13, unemployment rate. Value examples: \text{[}8.2, 7.9\text{]}.), \\
\text{\ \ \ \ }  (A15, number of crimes. Value examples: \text{[}256, 189\text{]}.) \\
\text{]} \\
\textbf{[Foreign keys]} \\
client.`district\_id` = district.`district\_id` \\
\textbf{[Question]} \\
What is the gender of the youngest client who opened account in the lowest average salary branch?
\textbf{[Evidence]} \\
Later birthdate refers to younger age; A11 refers to average salary \\
\textbf{[Answer]} \\
'''json \\
\{ \\
\text{\ \ \ \ } "account": "keep\_all", \\
\text{\ \ \ \ } "client": "keep\_all", \\
\text{\ \ \ \ } "loan": "drop\_all", \\
\text{\ \ \ \ } "district": \text{[}"district\_id", "A11", "A2", "A4", "A6", "A7"\text{]} \\
\} \\
''' \\
Question Solved. \\
==========  \\
 \\
Here is a new example, please start answering: \\

[DB\_ID] \{db\_id\} \\
\textbf{[Schema]} \\
\{desc\_str\} \\
\textbf{[Foreign keys]} \\
\{fk\_str\} \\
\textbf{[Question]} \\
\{query\} \\
\textbf{[Evidence]} \\
\{evidence\} \\
\textbf{[Answer]} \\
\endgroup
\end{dialogbox}

\subsection{Decomposer Prompt}
\label{appendix-decomposer-prompt}

\begin{dialogbox}
\begingroup
Given a \textbf{[Database schema]} description, a knowledge \textbf{[Evidence]} and the \textbf{[Question]}, you need to use valid SQLite and understand the database and knowledge, and then decompose the question into subquestions for text-to-SQL generation. \\
\\
When generating SQL, we should always consider constraints: \\
\textbf{[Constraints]} \\
- In `SELECT <column>`, just select needed columns in the [Question] without any unnecessary column or value  \\
- In `FROM <table>` or `JOIN <table>`, do not include unnecessary table  \\
- If use max or min func, `JOIN <table>` FIRST, THEN use `SELECT MAX(<column>)` or `SELECT MIN(<column>)`  \\
- If [Value examples] of <column> has 'None' or None, use `JOIN <table>` or `WHERE <column> is NOT NULL` is better  \\
- If use `ORDER BY <column> ASC|DESC`, add `GROUP BY <column>` before to select distinct values  \\
  \\
==========  \\

\textbf{[Database schema]} \\
\# Table: frpm \\
\text{[} \\
\text{\ \ \ \ } (CDSCode, CDSCode. Value examples: \text{[}'01100170109835', '01100170112607'\text{]}.), \\
\text{\ \ \ \ } (Charter School (Y/N), Charter School (Y/N). Value examples: \text{[}1, 0, None\text{]}. And 0: N;. 1: Y), \\
\text{\ \ \ \ } (Enrollment (Ages 5-17), Enrollment (Ages 5-17). Value examples: \text{[}5271.0, 4734.0\text{]}.), \\
\text{\ \ \ \ } (Free Meal Count (Ages 5-17), Free Meal Count (Ages 5-17). Value examples: \text{[}3864.0, 2637.0\text{]}. And eligible free rate = Free Meal Count / Enrollment) \\
\text{]} \\
\# Table: satscores \\
\text{[} \\
\text{\ \ \ \ } (cds, California Department Schools. Value examples: \text{[}'10101080000000', '10101080109991'\text{]}.), \\
\text{\ \ \ \ } (sname, school name. Value examples: \text{[}'None', 'Middle College High', 'John F. Kennedy High', 'Independence High', 'Foothill High'\text{]}.), \\
\text{\ \ \ \ } (NumTstTakr, Number of Test Takers in this school. Value examples: \text{[}24305, 4942, 1, 0, 280\text{]}. And number of test takers in each school), \\
\text{\ \ \ \ } (AvgScrMath, average scores in Math. Value examples: \text{[}699, 698, 289, None, 492\text{]}. And average scores in Math),
\text{\ \ \ \ } (NumGE1500, Number of Test Takers Whose Total SAT Scores Are Greater or Equal to 1500. Value examples: \text{[}5837, 2125, 0, None, 191\text{]}. And Number of Test Takers Whose Total SAT Scores Are Greater or Equal to 1500. . commonsense evidence:. . Excellence Rate = NumGE1500 / NumTstTakr) \\
\text{]} \\
\textbf{[Foreign keys]} \\
frpm.`CDSCode` = satscores.`cds` \\
\textbf{[Question]} \\
List school names of charter schools with an SAT excellence rate over the average. \\
\textbf{[Evidence]} \\
Charter schools refers to `Charter School (Y/N)` = 1 in the table frpm; Excellence rate = NumGE1500 / NumTstTakr \\

Decompose the question into sub questions, considering \textbf{[Constraints]}, and generate the SQL after thinking step by step: \\
Sub question 1: Get the average value of SAT excellence rate of charter schools. \\
SQL \\
''' \text{\ }sql \\
SELECT AVG(CAST(T2.`NumGE1500` AS REAL) / T2.`NumTstTakr`) \\
\text{\ \ \ \ } FROM frpm AS T1 \\
\text{\ \ \ \ } INNER JOIN satscores AS T2 \\
\text{\ \ \ \ } ON T1.`CDSCode` = T2.`cds` \\
\text{\ \ \ \ } WHERE T1.`Charter School (Y/N)` = 1 \\
''' \text{\ } \\ \\

Sub question 2: List out school names of charter schools with an SAT excellence rate over the average. \\
SQL \\
''' \text{\ }sql \\
SELECT T2.`sname` \\
\text{\ \ \ \ } FROM frpm AS T1 \\
\text{\ \ \ \ } INNER JOIN satscores AS T2 \\
\text{\ \ \ \ } ON T1.`CDSCode` = T2.`cds` \\
\text{\ \ \ \ } WHERE T2.`sname` IS NOT NULL \\
\text{\ \ \ \ } AND T1.`Charter School (Y/N)` = 1 \\
\text{\ \ \ \ } AND CAST(T2.`NumGE1500` AS REAL) / T2.`NumTstTakr` > ( \\
\text{\ \ \ \ } \text{\ \ \ \ } SELECT AVG(CAST(T4.`NumGE1500` AS REAL) / T4.`NumTstTakr`) \\
\text{\ \ \ \ } \text{\ \ \ \ } FROM frpm AS T3 \\
\text{\ \ \ \ } \text{\ \ \ \ } INNER JOIN satscores AS T4 \\
\text{\ \ \ \ } \text{\ \ \ \ } ON T3.`CDSCode` = T4.`cds` \\
\text{\ \ \ \ } \text{\ \ \ \ } WHERE T3.`Charter School (Y/N)` = 1 \\
\text{\ \ \ \ } ) \\
''' \text{\ } \\
 \\
Question Solved. \\
 \\
========== \\
 \\
\textbf{[Database schema]} \\
\# Table: account \\
\text{[} \\
\text{\ \ \ \ } (account\_id, the id of the account. Value examples: \text{[}11382, 11362, 2, 1, 2367\text{]}.), \\
\text{\ \ \ \ } (district\_id, location of branch. Value examples: \text{[}77, 76, 2, 1, 39\text{]}.), \\
\text{\ \ \ \ } (frequency, frequency of the acount. Value examples: \text{[}'POPLATEK MESICNE', 'POPLATEK TYDNE', 'POPLATEK PO OBRATU'\text{]}.), \\
\text{\ \ \ \ } (date, the creation date of the account. Value examples: \text{[}'1997-12-29', '1997-12-28'\text{]}.) \\
\text{]} \\
\# Table: client \\
\text{[} \\
\text{\ \ \ \ } (client\_id, the unique number. Value examples: \text{[}13998, 13971, 2, 1, 2839\text{]}.), \\
\text{\ \ \ \ } (gender, gender. Value examples: \text{[}'M', 'F'\text{]}. And F:female . M:male ), \\
\text{\ \ \ \ } (birth\_date, birth date. Value examples: \text{[}'1987-09-27', '1986-08-13'\text{]}.), \\
\text{\ \ \ \ } (district\_id, location of branch. Value examples: \text{[}77, 76, 2, 1, 39\text{]}.) \\
\text{]} \\
\# Table: district \\
\text{[} \\
\text{\ \ \ \ } (district\_id, location of branch. Value examples: \text{[}77, 76, 2, 1, 39\text{]}.), \\
\text{\ \ \ \ } (A4, number of inhabitants . Value examples: \text{[}'95907', '95616', '94812'\text{]}.), \\
\text{\ \ \ \ } (A11, average salary. Value examples: \text{[}12541, 11277, 8114\text{]}.)
\text{]} \\
\textbf{[Foreign keys]} \\
account.`district\_id` = district.`district\_id` \\
client.`district\_id` = district.`district\_id` \\
\textbf{[Question]} \\
What is the gender of the youngest client who opened account in the lowest average salary branch? \\
\textbf{[Evidence]} \\
Later birthdate refers to younger age; A11 refers to average salary \\

Decompose the question into sub questions, considering [Constraints], and generate the SQL after thinking step by step: \\
Sub question 1: What is the district\_id of the branch with the lowest average salary? \\
SQL \\
''' \text{\ }sql \\
SELECT `district\_id` \\
\text{\ \ \ \ } FROM district \\
\text{\ \ \ \ } ORDER BY `A11` ASC \\
\text{\ \ \ \ } LIMIT 1 \\
''' \text{\ } \\
 \\
Sub question 2: What is the youngest client who opened account in the lowest average salary branch? \\
SQL \\
''' \text{\ }sql \\
SELECT T1.`client\_id` \\
\text{\ \ \ \ } FROM client AS T1 \\
\text{\ \ \ \ } INNER JOIN district AS T2 \\
\text{\ \ \ \ } ON T1.`district\_id` = T2.`district\_id` \\
\text{\ \ \ \ } ORDER BY T2.`A11` ASC, T1.`birth\_date` DESC  \\
\text{\ \ \ \ } LIMIT 1 \\
''' \text{\ } \\  \\

Sub question 3: What is the gender of the youngest client who opened account in the lowest average salary branch? \\
SQL \\
''' \text{\ }sql \\
SELECT T1.`gender` \\
\text{\ \ \ \ } FROM client AS T1 \\
\text{\ \ \ \ } INNER JOIN district AS T2 \\
\text{\ \ \ \ } ON T1.`district\_id` = T2.`district\_id` \\
\text{\ \ \ \ } ORDER BY T2.`A11` ASC, T1.`birth\_date` DESC  \\
\text{\ \ \ \ } LIMIT 1  \\
''' \text{\ } \\
Question Solved. \\  \\

========== \\  \\

\textbf{[Database schema]} \\
\{desc\_str\} \\
\textbf{[Foreign keys]} \\
\{fk\_str\} \\
\textbf{[Question]} \\
\{query\} \\
\textbf{[Evidence]} \\
\{evidence\} \\  \\

Decompose the question into sub questions, considering [Constraints], and generate the SQL after thinking step by step: \\

\endgroup
\end{dialogbox}

\subsection{Refiner Prompt}

\begin{dialogbox}
\begingroup

\textbf{[Instruction]} \\
When executing SQL below, some errors occurred, please fix up SQL based on query and database info. Solve the task step by step if you need to. Using SQL format in the code block, and indicate script type in the code block. When you find an answer, verify the answer carefully. Include verifiable evidence in your response if possible. \\
\textbf{[Constraints]}  \\
- In `SELECT <column>`, just select needed columns in the \textbf{[Question]} without any unnecessary column or value  \\
- In `FROM <table>` or `JOIN <table>`, do not include unnecessary table  \\
- If use max or min func, `JOIN <table>` FIRST, THEN use `SELECT MAX(<column>)` or `SELECT MIN(<column>)`  \\
- If [Value examples] of <column> has 'None' or None, use `JOIN <table>` or `WHERE <column> is NOT NULL` is better  \\
- If use `ORDER BY <column> ASC|DESC`, add `GROUP BY <column>` before to select distinct values  \\
\textbf{[Query]}   \\
\{query\}   \\
\textbf{[Evidence]}   \\
\{evidence\}   \\
\textbf{[Database info]}   \\
\{desc\_str\}   \\
\textbf{[Foreign keys]}   \\
\{fk\_str\}   \\
\textbf{[old SQL]}   \\
''' \text{\ }sql   \\
\{sql\}   \\
''' \text{\ }   \\
\textbf{[SQLite error]}    \\
\{sqlite\_error\}   \\
\textbf{[Exception class]}   \\
\{exception\_class\}   \\
   \\
Now please fixup old SQL and generate new SQL again.   \\
\textbf{[correct SQL] }  \\

\endgroup
\end{dialogbox}

\section{Error Type Examples}
\label{sec:error-examples}

\begin{figure*}[]
    \centering
    \includegraphics[width=0.99\textwidth]{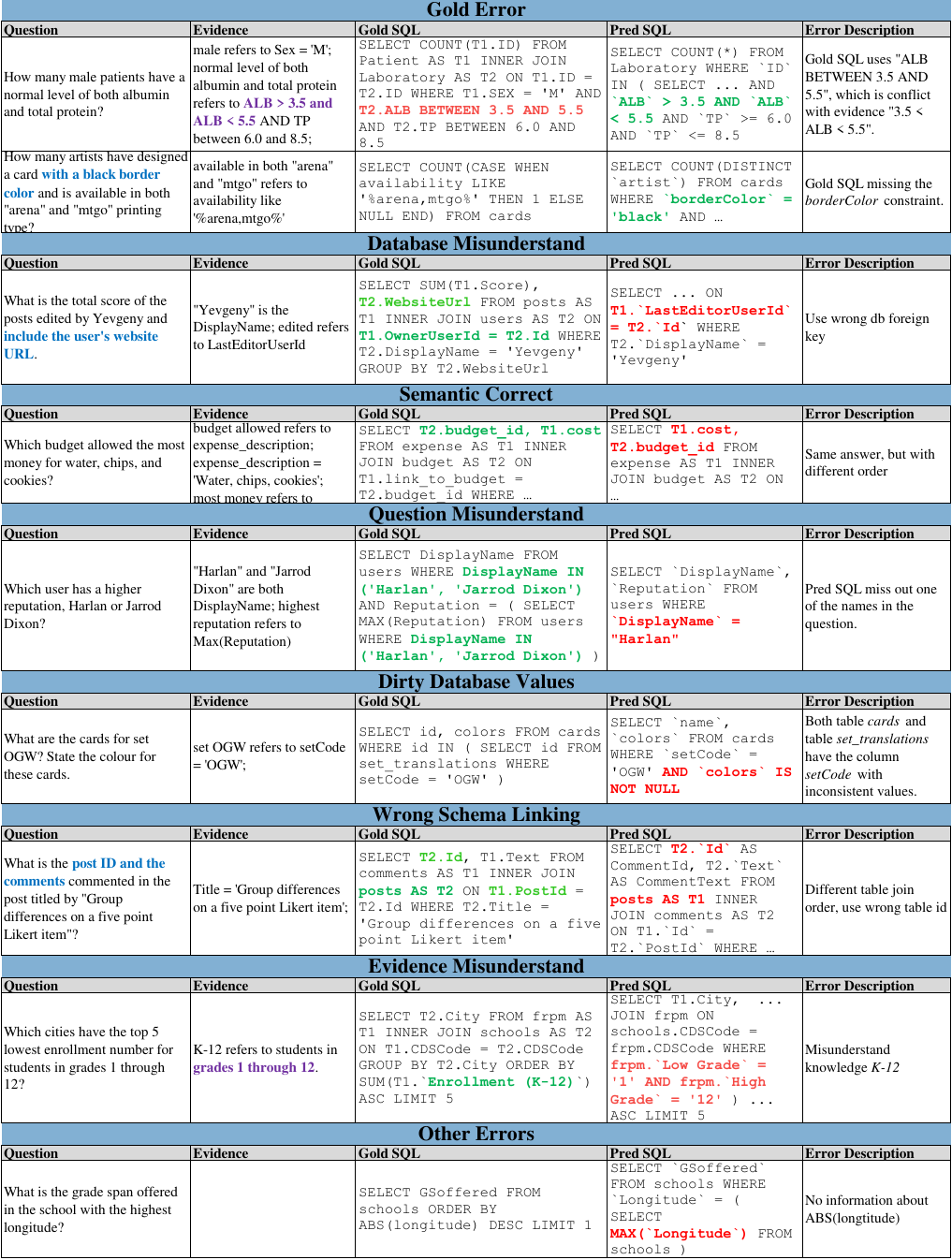}
    \caption{8 major types of error cases of BIRD are presented. Some cases are shortcuts for better presentation.}
    \label{fig:error_cases}
\end{figure*}

Examples of error types can be observed in Figure~\ref{fig:error_cases} (next page).

\end{document}